\documentclass[a4paper,fleqn]{cas-dc}

\usepackage[numbers]{natbib}
\hypersetup{colorlinks=true,linkcolor=red}
\usepackage{amsmath}
\usepackage{amssymb}
\usepackage{mathtools}
\usepackage[]{color,soul}
\usepackage{enumitem}

\def\tsc#1{\csdef{#1}{\textsc{\lowercase{#1}}\xspace}}
\tsc{WGM}
\tsc{QE}
\begin{document}
\let\WriteBookmarks\relax
\def\floatpagepagefraction{1}
\def\textpagefraction{.001}

\shorttitle{Improving DNN Classification Confidence using Heatmap-based XAI}    

\shortauthors{E. Tjoa, Guan C.}  

\title [mode = title]{Improving Deep Neural Network Classification Confidence using Heatmap-based eXplainable AI}  

%\tnotemark[<tnote number>] 

%\tnotetext[<tnote number>]{<tnote text>} 


\author[]{Erico Tjoa}[orcid=0000-0002-1599-1594]
%% Corresponding author indication
%\cormark[<corr mark no>]
%
%% Footnote of the first author
\fnmark[1]
%
%% Email id of the first author
%\ead{<email address>}
%
%% URL of the first author
%\ead[url]{<URL>}
%
%% Credit authorship
%% eg: \credit{Conceptualization of this study, Methodology, Software}
%\credit{<Credit authorship details>}
%
%% Address/affiliation
\affiliation[]{organization={Nanyang Technological University},
            addressline={50 Nanyang Ave}, 
            city={Singapore},
%          citysep={}, % Uncomment if no comma needed between city and postcode
            postcode={639798}, 
            state={},
            country={Singapore}}

\author[]{ Hong Jing Khok}\fnmark[1]
\author[]{Tushar Chouhan}
\fnmark[2]
            
\author[]{Guan Cuntai}[orcid=0000-0002-0872-3276]
%
%% Footnote of the second author
\fnmark[2]

%
%% Email id of the second author
%\ead{}
%
%% URL of the second author
%% Credit authorship
%% Address/affiliation

%% Corresponding author text
%\cortext[1]{Corresponding author}
%
%% Footnote text
\fntext[1]{Also affiliated with Alibaba Inc.}
\fntext[2]{School of Computer Science and Engineering (SCSE), NTU}


\begin{abstract}
This paper quantifies the quality of heatmap-based eXplainable AI (XAI) methods w.r.t image classification problem. Here, a heatmap is considered desirable if it improves the probability of predicting the correct classes. Different XAI heatmap-based methods are empirically shown to improve classification confidence to different extents depending on the datasets, e.g.  Saliency works best on ImageNet and Deconvolution on Chest X-Ray Pneumonia dataset. The novelty includes a new gap distribution that shows a stark difference between correct and wrong predictions. Finally, the generative augmentative explanation is introduced, a method to generate heatmaps capable of improving predictive confidence to a high level.
\end{abstract}


%% Research highlights
%\item 
%\item 
%\item 

% Keywords

\maketitle

% Main text
%\section{}\label{}
\section{Introduction}

Artificial intelligence (AI) and machine learning (ML) models have been developed with various levels of transparency and interpretability. Recent issues related to the responsible usage of AI have been highlighted by large companies like Google \cite{lakgoogle} and Meta \cite{pesenti}; this may reflect the increasing demand for transparency and interpretability, hence the demand for eXplainable Artificial Intelligence (XAI). In particular, the blackbox nature of a deep neural network (DNN) is a well-known problem in XAI. Many attempts to tackle the problem can be found in surveys like \cite{8466590,8400040,8631448,9233366}. 

Popular XAI methods include post-hoc methods such as Local Interpretable Model-agnostic Explanations (LIME) \cite{10.1145/2939672.2939778} and SHapley Additive exPlanations (SHAP) that uses a game-theoretical concept \cite{NIPS2017_7062}. Many heatmap-generating XAI methods have also been developed for DNN, in particular Class Activation Mappings (CAM) \cite{CAM7780688,DBLP:journals/corr/SelvarajuDVCPB16}, Layerwise Relevance Propagation (LRP) \cite{10.1371/journal.pone.0130140} and many other well-known methods, as listed in aforementioned surveys papers. These methods are appealing because heatmap-like attributions are intuitive and easy to understand. Although there are other remarkable ways to investigate interpretability and explainability e.g. methods that directly attempt to visualize the inner working of a DNN \cite{10.1007/978-3-319-10590-1_53,Olah_2017, olah_buildingblock}, we do not cover them here. This paper focuses on heatmap-based methods.

\textbf{How good are heatmap-based XAI methods}. Several existing efforts have also been dedicated to quantitatively measure the quality of heatmaps and other explanations. For example, heatmaps have been measured by their potentials to improve object localization performance \cite{CAM7780688,DBLP:journals/corr/SelvarajuDVCPB16}. The \textit{pointing game} \cite{8237633,9157775} is another example where localization concept is used to quantify XAI's performance. The ``most relevant first" (MORF) framework has also been introduced to quantify the explainability of heatmaps by ordered removal of pixels based on their importance \cite{7552539}; the MORF paper also emphasizes that there is a difference between \textit{computational relevance} and \textit{human relevance} i.e. objects which algorithms find salient may not be necessarily salient for a human observer. Others can be found e.g. in \cite{Tjoa2020QuantifyingEO}. \textit{This paper quantifies the quality of a heatmap based on how much the heatmap improves classification confidence.}

\textbf{Using heatmaps to improve the classification confidence of DNN}.  Heatmaps have been said to not ``[tell] us anything except where the network is looking at" \cite{Rudin2019}. In this work, we would like to refute such claims and show that heatmaps can be computationally useful. To test the usefulness of heatmaps in a direct way, we perform the \textit{Augmentative eXplanation (AX)} process: combine an image \(x\) with its heatmap \(h\) to obtain higher probability of predicting the correct class, e.g. if \(f(x)\) gives a \(60\%\) probability of making a correct prediction, we consider using \(h\) such that \(f(x+h)\) yields \(65\%\). We empirically show  existing XAI methods have the potential to improve classification confidence. However, heatmaps are usually not designed to explicitly improve classification performance, hence improvements are not observed in general.  This improvement is quantified through a metric we call the \textit{Confidence Optimization} (CO) score. Briefly speaking, CO score is a weighted difference between raw output values before and after heatmaps/attributions modify the images \(x+h\). The metric assigns a positive/negative score if \(x+h\) increases/decreases the probability of making the correct prediction. 

This paper is arranged as the following. In the next section, AX and Generative AX (GAX) are demonstrated through a two-dimensional toy example. Explicit form of heatmaps/attribution values can be obtained in the toy example, useful for lower level analysis and direct observation. The following section describes dataset pre-processing, computation of CO scores for AX process on existing XAI methods, formal definition GAX process and the results. We then present our results, starting with the novel finding: distribution \textit{gap} as correctness indicators, CO scores distribution for common XAI methods, followed by high scores attained by GAX heatmaps and finally qualitative aspects of the methods. Our results on ImageNet and Chest X-Ray Images for Pneumonia detection will be presented in the main text. Similar results on other datasets (1) COVID-19 Radiography Database (2) credit card fraud detection (3) dry bean classification can be found in the appendix. All codes are available in \url{https://github.com/ericotjo001/explainable_ai/tree/master/gax}.

\section{Formulation in Low Dimensional Space}

\begin{figure}[ht]
\vskip 0.2in
\begin{center}
\centerline{\includegraphics[width=\columnwidth]{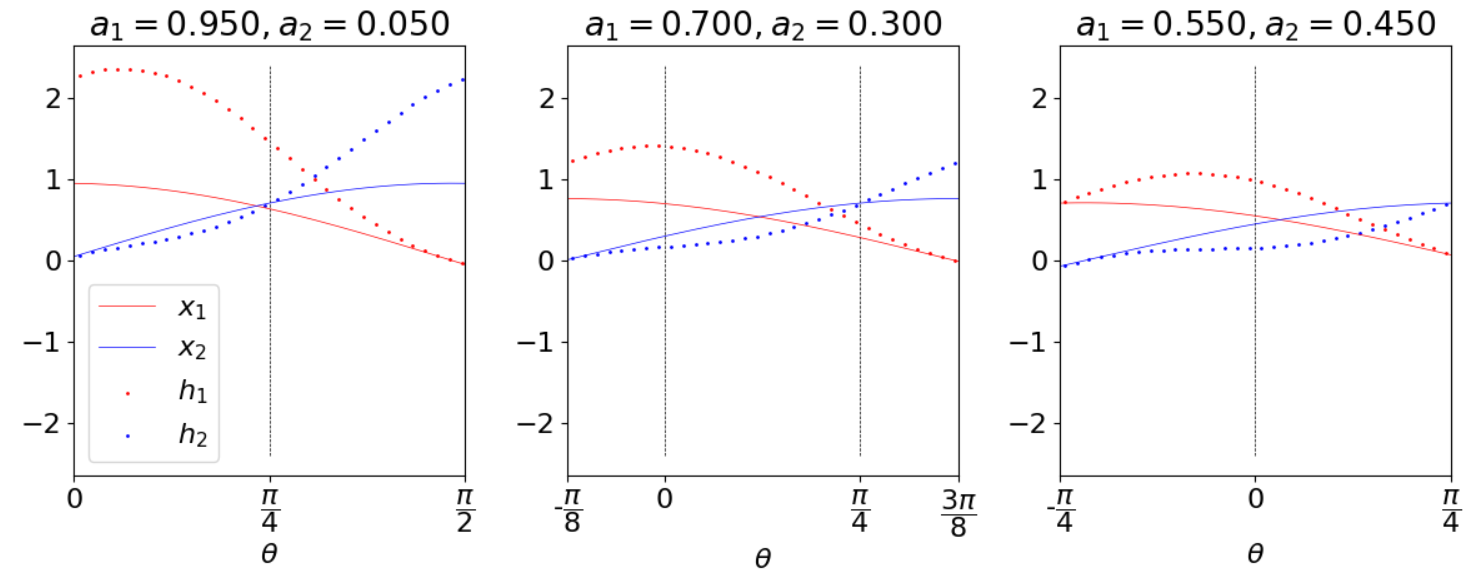}}
\caption{Solid red (blue) lines are \(x_1(x_2)\) components of sample data \(x\). Dotted red (blue) lines are \(h_1(h_2)\) components of heatmaps \(h\) with \(k\eta=1.2\). Heatmap values or attribute importances are assigned large values when either (1) the true components \(a_1,a_2\) differ significantly (2) the \(W\) transforms the data heterogenously i.e. not \(\theta\approx (2k+1)\frac{\pi}{4}\). See \textit{interpretations} in the main text for more details.}
\label{fig:rotate}
\end{center}
\vskip -0.2in
\end{figure}

The application presented in this paper is based on the following concept. We illustrate the idea using binary classification of data sample \(x\in\mathbb{R}^2\), a 2D toy example. Let \(y=W^{-1}x\) where \(y\in\mathbb{R}^2\) and \(W\in\mathbb{R}^{2\times 2}\) is invertible. Let the true label/category of sample \(x\) be \(c=argmax_{i}y_i\) so that it is compatible with one-hot encoding usually used in a DNN classification task. Conventions:
\begin{enumerate}[leftmargin=*,topsep=0pt]
\item \textit{Output space \(Y\)}. Let the output variable be \(y=a_1 \begin{psmallmatrix}\\ 1 \\ 0 \end{psmallmatrix} + a_2 \begin{psmallmatrix}\\ 0 \\ 1 \end{psmallmatrix}\), clearly an element of a vector space. The shape of this vector is the same as the output shape of the last fully connected layer for the standard binary classification. Class prediction can be performed in the winner-takes-all manner, for example, if \(a_1=1,a_2=0\), then the label is \(c=argmax_{i}y_i=1\). If \(a_1=0.1,a_2=0.5\), then \(c=2\). \textit{Basis of \(Y\)} is \(B_Y=\big\{ y^{(1)}=\begin{psmallmatrix}\\ 1 \\ 0 \end{psmallmatrix}, y^{(2)}=\begin{psmallmatrix}\\ 0 \\ 1 \end{psmallmatrix} \big\}\).
\item \textit{Sample space \(X\)} is a vector space with the corresponding basis \(B_X=\{Wy: y\in B_Y\}=\{x^{(1)}=Wy^{(1)},x^{(2)}=Wy^{(2)}\}\) so \(x=a_1 x^{(1)}+a_2 x^{(2)}\in X\). 
\item \textit{Pixelwise sample space} is the same sample space, but we specifically distinguish it as the sample space with the canonical basis. We will need this later, because pixelwise space has ``human relevance", since human observers perceive the components (pixels) directly, rather than automatically knowing the underlying structure (i.e. we cannot see \(a_1,a_2\) directly). We denote a sample in this basis with \(x=x_1\begin{psmallmatrix}\\ 1 \\ 0 \end{psmallmatrix} + x_2 \begin{psmallmatrix} 0 \\1 \end{psmallmatrix}\). 
\item A heatmap or attribute vector \(h\) in this paper has the same shape as \(x\) and can be operated directly with \(x\) via component-wise addition. Thus, they also belong to sample space or the pixelwise sample space. Writing a heatmap in the sample space \(h=Ax^{(1)}+Bx^{(2)}\) is useful for obtaining a closed form expression later. 
\end{enumerate}
\hfill

\textbf{The perfect classifier, \(f\)}. Define \(f(x,\Theta)=\sigma(\Theta x)\) as a trainable classifier with parameters \(\Theta\in\mathbb{R}^{2\times 2}\). Let \(\Theta=W^{-1}\) and  the activation \(\sigma\) be any strictly monotonic function, like the sigmoid function. Then, the classifier \(f(x)=\sigma(W^{-1}x)\in\mathbb{R}^2\) is perfect, in the sense that, if \(a_1>a_2\), then \(c=argmax_i f_i(x)=1\); likewise if \(a_1<a_2\), then \(c=2\) and, for \(a_1=a_2\) either decision is equally probable. This is easily seen as the following: \(f(x)=\sigma(W^{-1}(a_1 x^{(1)}+a_2 x^{(2)}))=\sigma(a_1 \begin{psmallmatrix}\\ 1 \\ 0 \end{psmallmatrix} + a_2 \begin{psmallmatrix}\\ 0 \\ 1 \end{psmallmatrix})=\begin{psmallmatrix}\\ \sigma(a_1) \\ \sigma(a_2) \end{psmallmatrix}\).

\textbf{Confidence optimization score} (CO score), \(s_{co}\). In this section, we show a simple explicit form of CO score for better illustration; in the experimental method section, formal definition will be given. The score increases if \(x+h\) leads to an improvement in the probability of correctly predicting label \(c\), hence the score's definition depends on the groundtruth label. Throughout this section, for illustration, we use \(x=a_1x^{(1)}+a_2 x^{(2)}\) with groundtruth label \(c=1\), i.e. \(a_1>a_2\). Define the CO score as
\begin{equation}
\label{eqn:COtoy}
s_{co}(x,h)=\begin{psmallmatrix} 1 \\ -1 \end{psmallmatrix}\cdot \big[f(x+h)-f(x)\big]
\end{equation}
For the perfect classifier, see that \(f_1(x+h)>f_1(x)\) and \(f_2(x+h)<f_2(x)\) contribute to a larger \(s_{co}\). In other words, increasing the probability of predicting the correct label \(c=1\) increases the score. For \(c=2\), replace \(\begin{psmallmatrix} 1 \\ -1 \end{psmallmatrix}\) with \(\begin{psmallmatrix} -1 \\ 1 \end{psmallmatrix}\).

\textbf{Augmentative explanation}. AX is defined here as any modification on \(x\) by \(h\) that is intended to yield positive the CO score, i.e to increase the probability of making a correct classification. This paper mainly considers the simplest implementation, namely \(x+h\). Let us consider a few possibilities. Suppose \(\sigma=LeakyReLU\) and \(h=x\). We get \(s_{co}=\begin{psmallmatrix} 1 \\ -1 \end{psmallmatrix}\cdot \begin{psmallmatrix}\\ \sigma(2a_1)-\sigma(a_1) \\ \sigma(2a_2)-\sigma(a_2) \end{psmallmatrix}=a_1-a_2>0\). In other words, choosing the image as the heatmap itself improves the score. However, as a heatmap or attribute vector, \(h\) is useless, since it does not provide us with any information about the relative importance of the components of \(x\) in canonical basis, which is the part of data directly visible to the observer. Even so, \(h=x\) has \textit{computational relevance} to the model, since \(a_1,a_2\) are modified in the correct direction. Our aim is to find computationally relevant \(h\) that does not score zero in ``human relevance", figuratively speaking. We therefore rule out obviously uninformative heatmap in the upcoming sections. Further, consider similar situation but set \(\sigma\) to sigmoid function. Simply setting \(h=x\) will no longer increase the score significantly all the time. Since sigmoid is \textit{asymptotic}, when \(a_1,a_2\) are sufficiently far away from zero, the increase will be so negligible, the heatmap will be uninformative even though the magnitude of \(|a_1-a_2|\) may be large. Hence, we use the raw DNN output in our main experiment, without sigmoid, softmax etc (in our experiment, we provide a comparison with the softmax version).

\textbf{Generative Augmentative EXplanation} (GAX) is an AX process where the heatmap \(h=w*x\) is generated by tuning the trainable parameter \(w\) so that  \(s_{CO}\) is optimized; \(*\) denotes component/pixel-wise multiplication. Here we will define \(\Delta=s_1\) as the term that we \textit{maximize} by hand, for clarity and illustration. By comparison, in the main experiment, we directly perform gradient descent on \(-s_1\) (plus regularization terms) to generate GAX heatmaps, i.e. we \textit{minimize} a total loss. To start with GAX, recall our choice of heatmap written in sample space basis,
\begin{equation}
\label{eqAB}
h=w* x=Ax^{(1)}+Bx^{(2)}
\end{equation}
This form is desirable as it can be manipulated more easily than the pixelwise sample space form \(h=\begin{psmallmatrix} w_1x_1\\w_2x_2  \end{psmallmatrix}\), as the following. From RHS of eq. (\ref{eqAB}), get \(AWy^{(1)}+BWy^{(2)}=W\begin{psmallmatrix}\\ A \\ B \end{psmallmatrix}\). We thus have \(\begin{psmallmatrix}A\\B\end{psmallmatrix}=W^{-1}(w*x)\). To increase CO score, the aim is to find parameter \(w\) that maximizes \(A-B\), i.e. find \(w^*=argmax_w(A-B)\).  Expanding the terms in \(w*x\) of eq. (\ref{eqAB}), we obtain \(\begin{psmallmatrix}w_1 \\ w_2 \end{psmallmatrix}* \begin{psmallmatrix} a_1W_{11}+a_2W_{12}\\ a_2W_{21}+a_2W_{22} \end{psmallmatrix}\). Taking the difference between the components gives us
\begin{equation}
\begin{aligned}
\Delta\equiv&A-B \\=& w_1(W_{11}^{-1}-W_{21}^{-1})(a_1W_{11} +a_2W_{12})\\
&-w_2(W_{22}^{-1}-W_{12}^{-1})(a_1W_{21}+a_2W_{22})
\end{aligned}
\end{equation}
Maximizing \(\Delta\) to a large \(\Delta>0\) will clearly optimize \(s_{co}(x, h)=\sigma(a_1)-\sigma(a_2)+\sigma(A)-\sigma(B)\), assuming \(\sigma\) is strictly monotonously increasing. 

\textit{Heatmap obtained through optimization using gradient ascent}. Recall that gradient ascent is done by \(\Delta\rightarrow \Delta+dw\cdot\nabla_w \Delta\) with the choice \(dw=\eta \nabla_w \Delta\), hence \(\Delta+\eta ||\nabla_w \Delta||^2\ge \Delta\). Hence, the heatmap after \(k\) steps of optimization is given by
\begin{equation}
\begin{aligned}
h=&(w+kdw)* x \\=&\big[w+k\eta \begin{psmallmatrix}(W_{11}^{-1}-W_{21}^{-1})(a_1W_{11} +a_2W_{12})\\-(W_{22}^{-1}-W_{12}^{-1})(a_1W_{21}+a_2W_{22})\end{psmallmatrix}\big]* x
\end{aligned}
\end{equation}
To visualize the heatmap, here we use the example where \(W\) is the rotation matrix \(W=\begin{psmallmatrix}cos\theta &-sin\theta\\sin\theta & cos\theta\end{psmallmatrix}\). Examples of heatmaps plotted along with the input \(x\) are shown in Figure \ref{fig:rotate}, to be discussed in the next subsection. If \(\theta=0\), \(x\) are identical to \(y\), so binary classification is straightforward and requires no explanation. Otherwise, consider \(\theta\) being a small deviation from \(0\). Such slightly rotated system is a good toy-example for the demonstration of component-wise ``importance attribution". This is because if \(x\) belongs to category \(c=1\) with high \(a_1\) component, then it still has a more significant first component \(x_1\) after the small rotation. Thus, a heatmap that correspondingly gives a higher score to the first component is ``correct" in the sense that it matches the intuition of attribute importance: high \(h_1\) emphasizes the fact that high \(x_1\) literally causes high \(y_1\). Furthermore, if the system rotates by \(\pi/4\), we see that the classification becomes harder. This is because the components \(x_1\) and \(x_2\) start to look more similar because \(cos \frac{\pi}{4}=sin \frac{\pi}{4}\), and consequently, the attribution values will be less prominent as well.

\subsection{Interpretability}
\textit{DISCLAIMER}: some reviewers who are familiar with a heatmap as \textit{the explanation} or a localization map tend to focus their attention on localization. While we do observe our resulting heatmaps, the formulas here are in no way aimed to improve object localization; we do not claim to achieve good localization. To reiterate, in this paper, heatmaps are evaluated based on their ability to computationally optimize the classification confidence.

\textit{Homogenous and Heterogenous transformations}. For the lack of better words, we refer to transformations like \(\theta\approx \pi/4\) or more generally \((2k+1)\frac{\pi}{4}\) for \(k=...,-1,0,1,...\) as homogenous transformations, since the components become more indistinguishable (recall: \(cos \frac{\pi}{4}=sin \frac{\pi}{4}\)). Otherwise, the transformation is called heterogenous. These definitions are given here with the intention of drawing parallels between (1) the toy data that have been homogenously transformed (hence hard to distinguish) and (2) samples in real datasets that look similar to each other, but are categorized differently due to a small, not obvious difference. 

\textit{Interpretation of attribute values for distinct non-negative components}. In the pixelwise sample space, we will be more interested in non-negative data sample \(x_1,x_2\ge 0\) since we only pass \([0,1]\)-normalized images for GAX. Figure \ref{fig:rotate} left shows a data sample with \textit{distinct} components, indicated by high \(a_1=0.95\) component and low \(a_2\). Non-negative data samples are found around \(\theta\in [0,\pi/2]\). High \(x_1\) value is given high \(h_1\) attribution score while low \(x_2\) is given a suppressed value of \(h_2\) near \(\theta=0\), matching our intuition as desired. As rotation proceeds to \(\pi/4\), there is a convergence between \(x_1\) and \(x_2\), making the components more indistinguishable. At \(\theta=\pi/4\) exactly, we still see high \(h_1\) that picks up high signal due to high \(a_1\), also as desired. Between \(\pi/4\) and \(\pi/2\), rotation starts to flip the components; in fact, at \(\pi/2\), \(x=[0,1]\) is categorized as \(c=1\) and \(x=[1,0]\) as \(c=2\). The attribution value \(h_2\) becomes more prominent, highlighting \(x_2\), also as desired for our prediction of class \(c=1\). In Figure \ref{fig:rotate} middle, decreased/increased \(a_1,a_2\) are assigned less prominent \(h_1,h_2\) respectively than Figure \ref{fig:rotate} left, since the model becomes less confident in its prediction, also consistent with our intuition.

\textit{The other extreme}. Figure \ref{fig:rotate} right shows \(a_1\) and \(a_2\) that do not differ significantly. At homogeneous transformation \(\theta\approx \pm\frac{\pi}{4}\), heatmaps are almost equal to the input \(x\). As expected, it will be difficult to pick up signals that are very similar, although very close inspection might reveal small differences that could probably yield some information (not in the scope of this paper). Other interpretations can be found in appendix \textit{More interpretations in low dimensional example}.

\section{Experimental Method and Datasets}
In the previous section, we described how heatmap \(h\) can be used to improve classification probability. More precisely, \(x+h\) yields higher confidence in making a correct prediction compared to \(x\) alone when used as the input to the model \(f\). We apply the same method to real dataset ImageNet \cite{deng2009imagenet} and Chest X-Ray Images (Pneumonia) from Kaggle \cite{mooney_2018}. The Pneumonia dataset needs reshuffling, since Kaggle's validation dataset consists of only of 16 images for healthy and pneumonia cases combined. We combined the training and validation datasets and then randomly draw 266/1083 healthy and 790/3093 pneumonia images for validation/training. There are 234/390 healthy/pneumonia images in the test dataset. Images are all resized to \(256\times 256\). The X-Ray images are black and white, so we stack them to 3 channels. Images from ImageNet are normalized according to suggestion in the pytorch website, with \(mean=[0.485, 0.456, 0.406], std=[0.229, 0.224, 0.225]\).

\begin{table}[t]
\caption{Fine-tuning results for pre-trained models on Chest X-Ray Pneumonia test dataset. The architectures marked with \textunderscore sub are deliberately trained to achieve lower validation accuracy for comparison.}
\label{table:finetune}
\vskip 0.15in
\begin{center}
\begin{small}
\begin{tabular}{c | c c c}
\toprule
& Resnet34\textunderscore 1 & Resnet34\textunderscore sub & Alexnet \textunderscore sub\\
\midrule
 accuracy & 0.800 & 0.636 & 0.745 \\  
 precision & 0.757 & 0.632 & 0.726 \\    
 recall & 1.000 & 1.000 & 0.951   \\
 val. acc. & 0.99 & 0.8 & 0.8\\
\bottomrule
\end{tabular}
\end{small}
\end{center}
\vskip -0.1in
\end{table}

\begin{figure*}[ht]
\vskip 0.2in
\begin{center}
\centerline{\includegraphics[width=2.1\columnwidth]{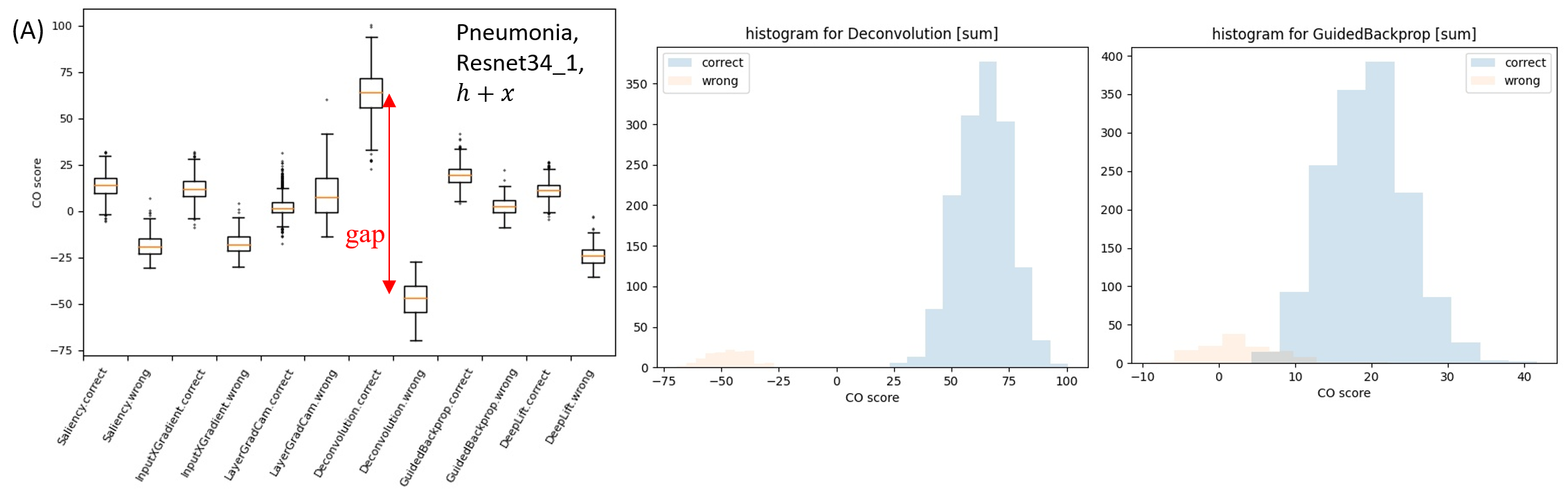}}
\caption{Distribution of CO scores obtained through AX process on existing XAI methods. Classification probability is improved if the score is positive. All distributions show gaps between CO scores of data whose classes are correctly and wrongly predicted (e.g. red arrows); correct prediction tends to yield higher CO scores. The result is obtained using Resnet34\textunderscore 1 on Pneumonia dataset. [sum] denotes AX process with \(x+h\).}
\label{fig:cogap}
\end{center}
\vskip -0.2in
\end{figure*}

For both datasets, we use pre-trained models Resnet34 \cite{He2016DeepRL} and AlexNet \cite{Krizhevsky2014OneWT} available in Pytorch. The models are used on ImageNet without fine-tuning. Resnet34 is fine-tuned for Pneumonia binary classifications, where the first 8 modules of the pre-trained model (according to pytorch's arrangement) are used, plus a new fully-connected (FC) layer with two output channels at the end. Similarly, for Alexnet, the first 6 modules are used with a two-channel FC at the end. For Resnet34, we will use Resnet34\textunderscore 1 and Resnet34\textunderscore  sub respectively trained to achieve \(99\%\) and \(80\%\) validation accuracies for comparison. The same targets were specified for Alexnet, but only \(80\%\) validation accuracy was achieved, thus only Alexnet\textunderscore sub will be used. Adam optimizer is used with learning rate \(0.001\), \(\beta=(0.5,0.999)\). The usual weight regularization is not used during optimization i.e. in pytorch's Adam optimizer, weight decay is set to zero because we allow zero attribution values in large patches of the images. No number of epochs are specified. Instead, training is stopped after the max number of iterations (240000) or the specified validation accuracy is achieved after 2400 iterations have passed. At each iteration, samples are drawn uniform-randomly with batch size 32.

Further details on COVID-19 Radiography Database, credit card fraud detection and dry bean classification can be found in the appendix.

\textbf{Multi-class CO scores on existing XAI methods via AX process}. Denote the deep neural network as DNN, define the CO score as the weighted difference between the predictive scores altered by AX process and the original predictive scores,
\begin{equation}
s_{co}(x,h)= \kappa\cdot \big[DNN(x+h)-DNN(x)\big] 
\end{equation}
where \(\kappa\in\mathbb{R}^C\) is defined as the \textit{score constants}, \(C\) the number of classes, \(\kappa_j=1\) if the groundtruth belongs to label/category \(j\) and \(\kappa_i=-1/(C-1)\) for all \(i\neq j\). This equation is the general form of eq. (\ref{eqn:COtoy}). In our implementation, each DNN's output is \textit{raw}, i.e. last layer is FC with \(C\) channels without softmax layer etc (the softmax version is available in the appendix). A heatmap \(h\) that yields \(s_{co}=0\) is uninformative (see appendix). We compute CO scores for heatmaps generated by six different existing heatmap-based XAI methods (all available in Pytorch Captum), namely, Saliency \cite{Simonyan14a}, Input*Gradient \cite{Shrikumar2016NotJA}, Layer GradCAM \cite{DBLP:journals/corr/SelvarajuDVCPB16}, Deconvolution \cite{10.1007/978-3-319-10590-1_53}, Guided Backpropagation \cite{Springenberg2015StrivingFS} and DeepLift \cite{pmlr-v70-shrikumar17a}. Each heatmap is generated w.r.t predicted target, not groundtruth e.g. if y\textunderscore pred=DNN(x) predicts class \(n\), then h=DeepLIFT(net).attribute(input, target=n) in Pytorch Captum notation. Then normalization is applied \(h\rightarrow h/max(|h|)\) before we perform the AX process. Note: For ImageNet, \(C=1000\), chest X-Ray, \(C=2\). We also consider \(f(x*h)\), where \(*\) denotes component-wise multiplication. The idea is generally to interact \(h\) with \(x\) so that, for any interaction \(g\), higher probability of correct prediction is achieved by \(f(g(x,h))\); see appendix for their results. GradCAM of `conv1' layer is used in this paper. Other methods and different arbitrary settings are to be tested in future works.

\textbf{Achieving high \(s_{co}\) with GAX}. Here, GAX is the \(x+h\) AX process where heatmaps \(h=tanh(w* x)\) are generated by training parameters \(w\) to maximize \(s_{co}\). Maximizing \(s_{co}\) indefinitely is impractical, and thus we have chosen \(s_{co}=48\) for ImageNet dataset, a score higher than most \(s_{co}\) attained by existing XAI methods we tested in this experiment.  Tanh activation is used both to ensure non-linearity and to ensure that the heatmap is normalized to \([-1,1]\) range, so that we can make a fair comparison with existing heatmap-based XAI methods. For ImageNet, 10000 data samples are randomly drawn from the validation dataset for evaluating GAX. For pneumonia, all data samples are used. Optimization is done with Adam optimizer with learning rate \(0.1\), \(\beta=(0.9,0.999)\). This typically takes less than 50 steps of optimization, a few seconds per data sample using a small GPU like NVIDIA GeForce GTX 1050.

\textbf{Similarity loss and GAX bias}. In our implementation, we minimize \(-s_{co}\). However, this is prone to producing heatmaps that are visually imperceptible from the image. Since \(w\) is initialized as an array of \(1\)s with exactly the same shape \((c,h,w)=(3,256,256)\) as \(x\), the initial heatmap is simply \(h=w*x=x\). Possibly, small changes in \(w\) over the entire pixel space is enough to cause large changes in the prediction, reminiscent of adversarial attack \cite{Szegedy2014IntriguingPO, 8294186}. We solve this problem by adding the similarity loss, penalizing \(h=x\). The optimization is now done by minimizing the modified loss, which is negative CO score plus \textit{similarity loss}
\begin{equation}
loss=-s_{co} + l_{s}\Big\langle \displaystyle\frac{(h-x+\epsilon)^2}{x+\epsilon}\Big\rangle^{-1}
\end{equation}
where \(l_{s}=100\) is the \textit{similarity loss factor}. \(\langle X\rangle\) computes the average over all pixels. Division \(/\) and square \(^2\) are performed component/pixel-wise. Pixel-wise division by \(x\) normalizes the pixel magnitude, so that small pixel values can contribute more significantly to the average value. The small term \(\epsilon=10^{-4}\) is to prevent division by zero and possibly helps optimization by ensuring that zero terms do not make the gradients vanish. Furthermore, for X-Ray images, with many zeros (black region), the similarity factor seems insufficient, resulting in heatmaps that mirror the input images. GAX bias is added for the optimization to work, so that \(h=w*x + b\), where \(b\) is \(0.01\) array of the same shape \((c,h,w)\) as well. Note: the similarity loss is positive, since \(x\) used here is \([0,1]\) normalized (by comparison, the standard Resnet34 normalization can result in negative pixels).

\begin{figure*}[ht]
\vskip 0.2in
\begin{center}
\centerline{\includegraphics[width=2.1\columnwidth]{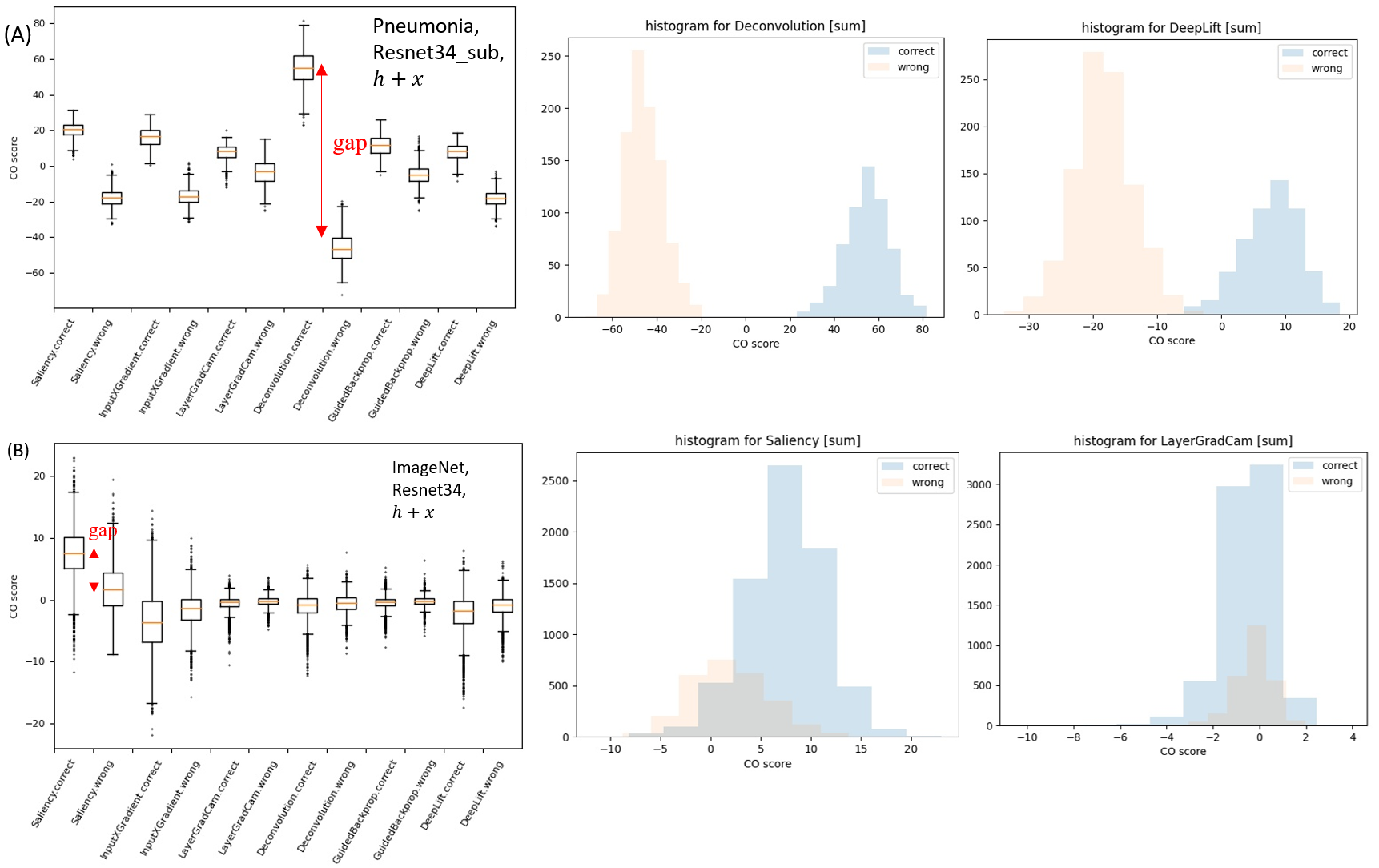}}
\caption{ Similar to Figure \ref{fig:cogap}, but the results are obtained from (A) Resnet34 architecture on Pneumonia dataset, but with less fine-tuning (Resnet34\textunderscore sub). (B) Resnet34 on ImageNet.}
\label{fig:cogap2}
\end{center}
\vskip -0.2in
\end{figure*}

\section{Results and Discussions}
Recall that we use pre-trained models for ImageNet. For pneumonia dataset, the predictive results of fine-tuning models are shown in table \ref{table:finetune}. AX and GAX processes will be applied on top of these models. Here, we observe possibly novel gaps in CO scores distribution and show that GAX can possibly be used to detect false distinct features. Similar results on COVID-19 Radiography Database, credit card fraud detection and dry bean classification will be presented in the appendix.

\subsection{Gaps in CO Scores Distribution}
Here, we present the main novel finding: the gap in CO distribution. AX process does not optimize any losses to distinguish correct predictions from the wrong ones, but Figure \ref{fig:cogap} shows distinct gaps between them (shown by the red arrows). Possible reason is as the following. Heatmaps used in AX process are generated for the class \(c_{pred}\) predicted by the DNN, e.g. h=DeepLIFT(net).attribute(input, target=\(c_{pred}\)); recall: we use Pytorch Captum notation. If the prediction is correct, there is a match between \(c_{pred}\) and the groundtruth label \(c\) that that affects CO score through \(\kappa\). 

\textit{What's the significance of the distribution gap?} The different distributions found in Figure \ref{fig:cogap} and \ref{fig:cogap2} indicate that some existing XAI methods possess more information to distinguish between correct and wrong predictions than the others. With this, we might be able to debunk some claims that heatmaps are not useful \cite{Rudin2019}: regardless of the subjective assessment of heatmap shapes, heatmaps might be relatively informative after some post-processing. In the absence of such information, we expect to see uniformly random distribution of scores. Since we have observed distinct distributional gaps on top of general difference in the statistics, we have shown that some heatmap-based XAI methods combined with CO score might be a new indicator to help support classification decision made by the particular DNN architecture.

\textit{What's its application?} In practice, when new samples are provided for clinical support, human data-labeling might still be needed, i.e. assign ground-truth \(c\). However, errors sometimes happen. AX method can be used to flag any such error by matching the CO score (check whether the score lies above or before the large gap).

\begin{figure}[ht]
\vskip 0.2in
\begin{center}
\centerline{\includegraphics[width=0.8\columnwidth]{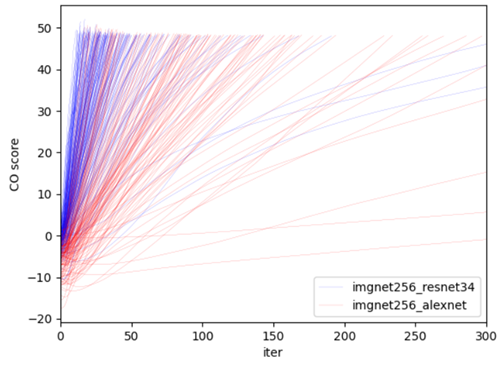}}
\caption{CO score optimization through GAX \(x+tanh(w*x)\) on ImageNet data using pre-trained Resnet34 (blue) and Alexnet (red), where each curve corresponds to a single image. The target \(s_{co}\) is set to 48, exceeding most CO scores of other methods.}
\label{fig:gax_op}
\end{center}
\vskip -0.2in
\end{figure}

\textit{Variation}. Furthermore, the extent of CO score distribution gap is clearly dependent on the dataset and DNN architecture. As it is, the discriminative capability of different XAI methods is thus comparable only within the same system of architecture and dataset.  ImageNet dataset shows a smaller gap compared to pneumonia dataset and the largest gap in ImageNet is produced by the Saliency method, as seen in Figure \ref{fig:cogap2}(B). By comparison, the largest gap in pneumonia dataset is produced by Deconvolution. Comparing Figure \ref{fig:cogap} and Figure \ref{fig:cogap2}(A), the gaps appear to be wider when DNN is better trained. Further investigation is necessary to explain the above observations, but, to leverage this property, users are encouraged to test AX process on different XAI methods to find the particular method that shows the largest gap. Once the XAI method is determined, it can be used as a supporting tool and indicator for the correctness of prediction.

\subsection{Improvement in Predictive Confidence with GAX}
The higher the CO scores are, the better is the improvement in predictive probability. Is it possible to achieve even higher improvement, i.e. higher CO score? Figure \ref{fig:cogap} and \ref{fig:cogap2} show the boxplots of CO scores for AX process applied on all six XAI methods we tested in this experiment; histograms applied on select XAI methods are also shown. Different XAI methods achieve different CO scores. For pneumonia dataset, very high CO scores (over 80) are attained by Deconvolution methods. For ImageNet, highest CO scores attained are around 10. To attain even higher scores, Generative AX (GAX) will be used.

\begin{figure*}[ht]
\vskip 0.2in
\begin{center}
\centerline{\includegraphics[width=2\columnwidth]{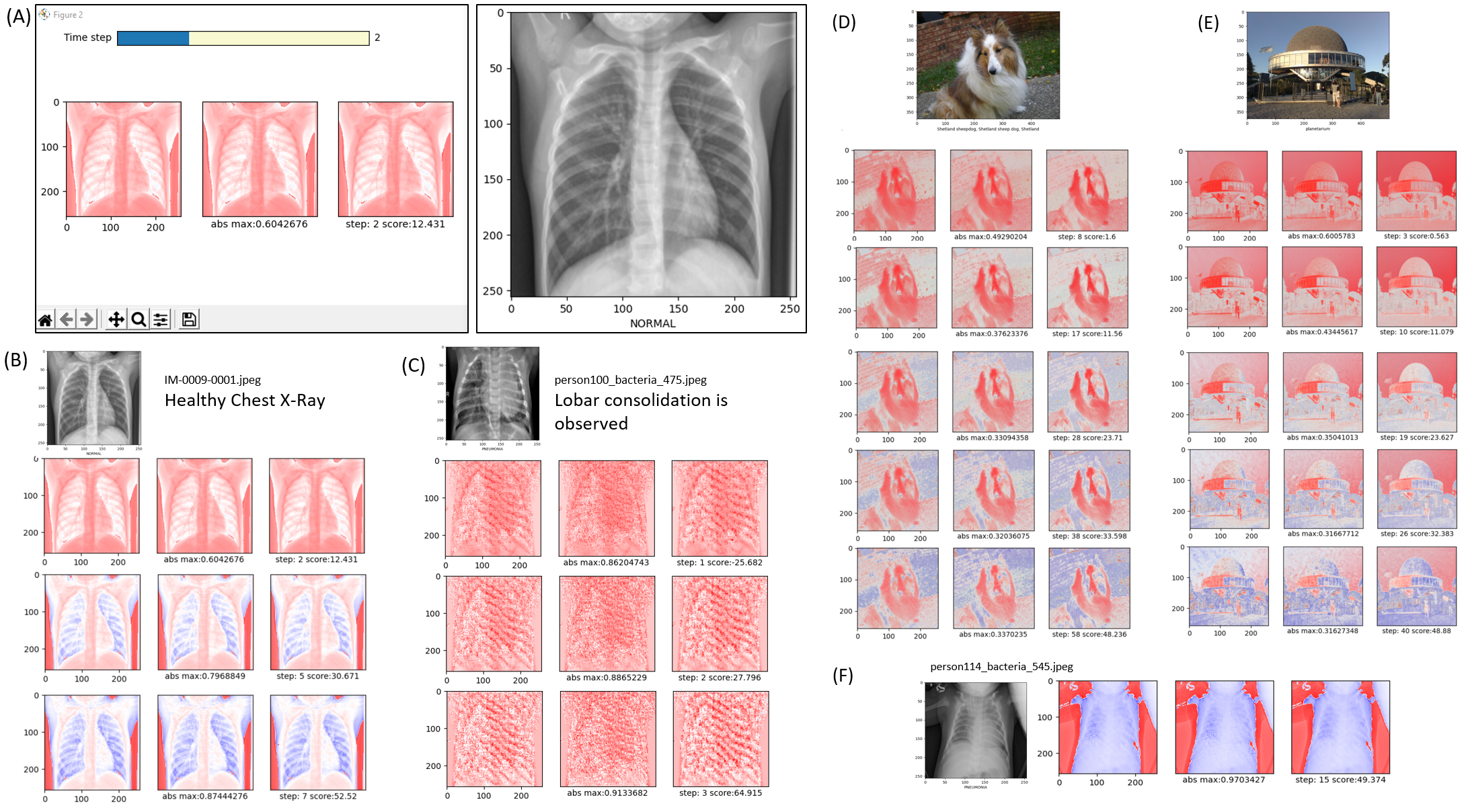}}
\caption{(A) GAX dynamic heatmaps displayed with a slider for users to observe the evolution of heatmaps through time steps. (B-E) GAX heatmaps generated on Resnet34 for (B) healthy chest X-Ray and (C) chest X-Ray of a patient with bacterial pneumonia; and for ImageNet images (D) a sheep dog image and (E) a planetarium image. (F) An instance of bacterial pneumonia chest X-Ray showing an irregular posture with heavy noise (top right). The empty space might have been used as a false distinct feature for pneumonia classification. Three heatmaps for each image correspond to the attribution values assigned to R, G and B color channels respectively. ``Abs max" specifies the maximum absolute value attained by the heatmap throughout all three channels (max is 1, due to Tanh activation). Positive/negative heatmap or attribution values (red/blue) indicate pixels to be increased/reduced in intensity to attain higher prediction confidence. At higher CO scores, negative values emerge.}
\label{fig:qualitative}
\end{center}
\vskip -0.2in
\end{figure*}

Using GAX on ImageNet, \(s_{co}\ge 48\) can be attained as shown in Figure \ref{fig:gax_op}, where the time evolution of CO score for each image is represented by a curve. For Resnet34, most of the images attains \(s_{co}\ge 48\) within 50 iterations. Alexnet GAX optimization generally takes more iterations to achieve the target. High \(s_{co}\) implies high confidence in making the correct prediction. We have thus obtained heatmaps and attribution values with computational relevance i.e. they can be used to improve the model's performance. Note: (1) all images tested do attain the target \(s_{co}\) (not shown), although some of the images took a few hundreds iterations (2) we exclude images where predictions are made incorrectly by the pre-trained or fine-tuned model. By comparison, in general, using heatmaps derived from existing methods for AX process does not yield positive CO scores i.e. does not improve predictive probability for the correct class (see especially Figure \ref{fig:cogap2}(B)). Furthermore, for ImageNet, typically, \(s_{co}\le 10\). Other boxplots are shown in appendix (including results run with pytorch v2.0).

The improvement of predictive accuracy across the entire samples within a particular dataset has been limited. As shown in table \ref{table:allacc} (appendix), we only observe small improvements for instances where the original performances are sub-optimal (coloured blue). To be fair, some of them can even be ruled out as noise. Some other methods retain their classification accuracy, which means they effectively improve the classification confidence at no cost, but some methods such as InputXGradient have degraded the accuracy across all datasets and models.

\subsection{Qualitative Assessment of GAX Heatmaps}
Heatmaps in GAX are obtained through a process optimization through a finite number of time steps. We provide matplotlib-based graphic user interface (GUI) for users to observe the evolution of heatmap pixels through GAX; see Figure \ref{fig:qualitative}(A). This provides users some information about the way the DNN architecture perceives input images. But, how exactly can user interpret this? Recall that the main premise of this paper is the computational relevance: GAX is designed to generate heatmaps that improve the confidence in predicting the correct label numerically. Hence, the visual cues generated by the GAX heatmaps show which pixels can be increased or decreased in intensity to give higher probability of making the correct prediction. 

\textit{DNN optimizes through extreme intensities}. Heatmaps in Figure \ref{fig:qualitative}(B-E) show that predictive confidence is improved generally through optimizing regions of extreme intensity. For example, to improve CO scores through GAX, the pixels corresponding to white hair of the dog in Figure \ref{fig:qualitative}(D) are assigned positive values (red regions in the heatmaps). Dark region of healthy chest X-Ray in (B) are subjected to stronger optimization (intense red or blue) to achieve better predictive confidence. The DNN architecture seems to be more sensitive to changes in extreme values in the image. In a positive note, this property might be exploited during training: this is probably why normalization to \([-1,1]\) range in the standard practice of deep learning optimization works compared to \([0,1]\). On the other hand, this might be a problem to address in the future as well: heatmaps that boost algorithmic confidence are not intuitive to human viewers. We can ask the question: is it possible to train DNN such that its internal structure is inherently explainable (e.g. if localization is accepted as an explanation, does there exist an architecture whose predictive confidence is tied directly to localization?). For comparison, existing XAI methods typically specify extra settings to obtain these explanations. Unfortunately, the settings can be arbitrary, e.g. GradCAM paper \cite{DBLP:journals/corr/SelvarajuDVCPB16} sets an arbitrary \(15\%\) threshold of max intensity for binarization. To obtain explanation with better integrity, the settings might need to be specified in context beforehand. In this paper, we do NOT address such arbitrary settings taylored to attain subjectively acceptable explanation or to maximize high IoU for bounding boxes. 

\textit{Discriminatory but unrefined patterns}. Pneumonia dataset consists of chest X-Ray of healthy patients and patients with several types of pneumonia with different recognizable patterns. Bacterial pneumonia has a focal lobar consolidation, while viral pneumonia has diffuse interstitial patterns; normal chest X-Ray has neither. This turns out to affect the shape of GAX heatmaps. Figure \ref{fig:qualitative}(B) shows a typical normal Chest X-Ray pattern. By comparison, Figure \ref{fig:qualitative}(C) shows a heatmap generated on bacterial pneumonia. In the latter, we see the drastic change in the heatmap features, especially high-intensity stripes around the lobar consolidation. There is a possibility that novel class-discriminatory patterns lie hidden within heatmaps generated by GAX. The heatmaps appear unrefined, but this might be related to the internal structures of the DNN architecture itself, as described in the following section.

\textbf{Limitations and Future works}. (1) Optimized regions prefer extreme intensities (very bright or very dark regions). The heatmaps in Figure \ref{fig:qualitative}(B-E) indicate that we are able to optimize predictive probability through relative intensity manipulation of pixel patterns that are not humanly intuitive. To truly capture variations in patterns and not rely heavily on large difference in intensity, a layer or module specifically designed to output very smooth representation might be helpful. Training might take longer, but we hypothesize that skewed optimization through extreme intensity can be prevented. (2) Some optimized features are rife with artifact-looking patterns. An immediate hypothesis that we can offer is the following. The internal structure of the DNN (the set of weights) is noisy, thus, even if features are properly captured, they are amplified through noisy channels, yielding artifacts. This is indicative of the instability of high dimensional neuron activations in a DNN, a sign of fragility against adversarial attack we previously mentioned. How should we address this? We need DNN that are robust against adversarial attack; fortunately, many researchers have indeed worked on this problem recently. (3) The regularity of data distribution is probably an important deciding factor in model training. In cases where the X-Ray images are not taken in a regular posture, the empty space can become a false ``distinct feature", as shown in Figure \ref{fig:qualitative}(F). While this may indicate a worrying trend in the flawed training of DNN or data preparation (hence a misguided application) we believe GAX can be used to detect such issue before deployments. Related future studies may be aimed at quantifying the effect of skewed distribution on the appearance of such ``false prediction" cases. (4) In the future, GAX can be tested on different layers of a DNN e.g. we can leverage the weakly supervised localization of CAM to attain visually sensible heatmaps. Also see appendix for more, e.g. implementation-specific limitations etc.

\section{Conclusion} 
We have investigated a method to use heatmap-based XAI methods to improve DNN's classification performance. The method itself is called the AX process, and the improvement is measured using a metric called the CO score. Some heatmaps can be directly used to improve model's prediction better than the others as seen by the boxplots of score distribution. The distribution of scores shows a novel gap distribution, an interesting feature that develops without any specific optimization. GAX is also introduced to explicitly attain high improvement in predictive performance or help detect issues. This work also debunks claims that heatmaps are not useful through the improvement of predictive confidence. We also give explanations on DNN behaviour consistent with the standard practice of deep learning training. From the results, we support the notion that computationally relevant features are not necessarily relevant to human. 

Summary of novelties and contributions: (1) CO scores provide empirical evidence for informative content of heatmaps (2) the distribution gap in CO scores may be a new indicator in predictive modelling (3) distinct (albeit unrefined) class-dependent patterns that emerge on GAX-generated heatmaps could be used as discriminative signals. Overall, we also provide insights into the DNN's behaviour.

\section{Acknowledgment}
This research was supported by Alibaba Group Holding Limited, DAMO Academy, Health-AI division under Alibaba-NTU Talent Program, Alibaba JRI. The program is the collaboration between Alibaba and Nanyang Technological university, Singapore. This work was also supported by the RIE2020 AME Programmatic Fund, Singapore (No. A20G8b0102).

% Use the style of numbering in square brackets.
% If nothing is used, default style will be taken.
%\begin{enumerate}[a)]
%\item 
%\item 
%\item 
%\end{enumerate}  

%\begin{itemize}
%\item 
%\item 
%\item 
%\end{itemize}  

%\begin{description}
%\item[]
%\item[] 
%\item[] 
%\end{description}  

% Figure
%\begin{figure}[<options>]
%	\centering
%		\includegraphics[<options>]{}
%	  \caption{}\label{fig1}
%\end{figure}

%
%\begin{table}[<options>]
%\caption{}\label{tbl1}
%\begin{tabular*}{\tblwidth}{@{}LL@{}}
%\toprule
%  &  \\ % Table header row
%\midrule
% & \\
% & \\
% & \\
% & \\
%\bottomrule
%\end{tabular*}
%\end{table}

% Uncomment and use as the case may be
%\begin{theorem} 
%\end{theorem}

% Uncomment and use as the case may be
%\begin{lemma} 
%\end{lemma}

 \appendix

\printcredits

\bibliographystyle{cas-model2-names}

\bibliography{gaxelse}

%% Biography
%% Here goes the biography details.
%% Here goes the biography details.

\appendix
\onecolumn

\section{Appendix}

All codes are available in the supplementary materials. All instructions to reproduce the results can be found in README.md, given as command line input, such as:
\begin{enumerate}
\item python main\_pneu.py --mode xai\_collect --model resnet34 --PROJECT\_ID pneu256n\_1 --method Saliency --split train --realtime\_print 1 --n\_debug 0
\item python main\_pneu.py --mode gax --PROJECT\_ID pneu256n\_1 --model resnet34 --label NORMAL --split test --first\_n\_correct 100 --target\_co 48 --gax\_learning\_rate 0.1
\end{enumerate}

\noindent The whole experiment can be run on small GPU like NVIDIA GeForce GTX 1050 with 4 GB dedicated memory.

The codes are run on Python 3.8.5. The only specialized library used is Pytorch (specifically torch==1.8.1+cu102, torchvision==0.9.1+cu102) and Pytorch Captum (captum==0.3.1). Other libraries are common python libraries.

\textbf{Regarding Captum}. We replace Pytorch Captum ``inplace relu" so that some attribution methods will work properly (see adjust\_for\_captum\_problem in model.py where applicable). 

We also manually edit non-full backward hooks in the source codes to prevent the gradient propagation issues. For example, from Windows, see Lib \textbackslash site-packages \textbackslash captum \textbackslash attr \textbackslash \_core \textbackslash guided\_backprop\_deconvnet.py, function def \_register\_hooks(self, module: Module). There is a need to change from hook = module. register\_backward\_hook(self.\_backward\_hook) to hook = module. register\_full\_backward\_hook(self.\_backward\_hook).

\subsection{More interpretations in low dimensional example}
\label{appd:interpretation}
\begin{figure}[]
\centering
\includegraphics[width=0.5\columnwidth]{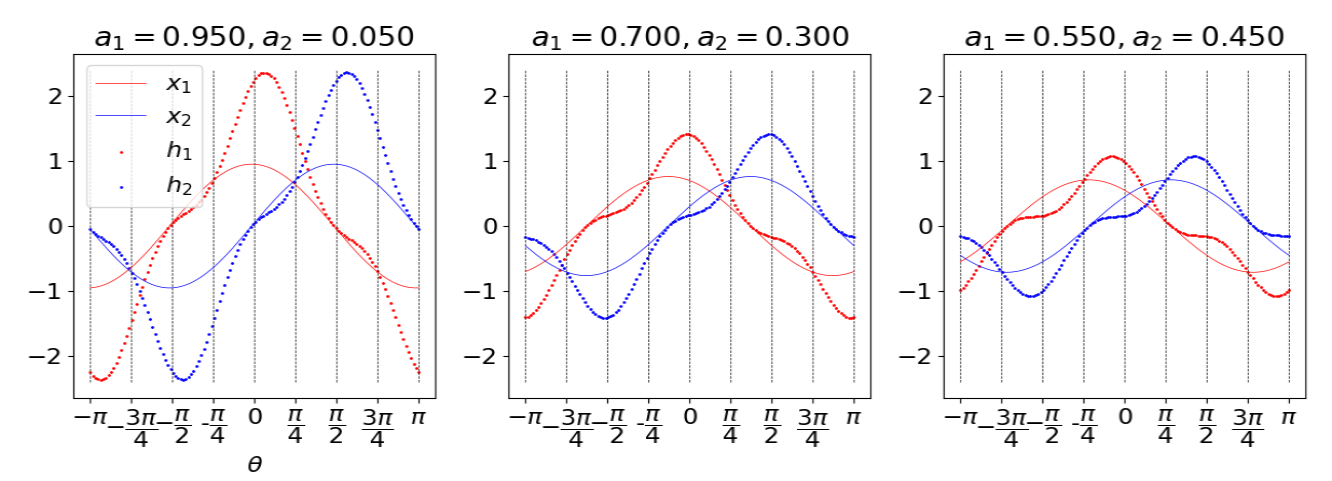} 
\caption{Solid red (blue) lines are \(x_1(x_2)\) components of sample data \(x\). Dotted red (blue) lines are \(h_1(h_2)\) components of heatmaps \(h\) with \(k\eta=1.2\). Heatmap values or attribute importances are assigned large values when either (1) the true components \(a_1,a_2\) differ significantly (2) the \(W\) transforms the data heterogenously i.e. not \(\theta\approx (2k+1)\frac{\pi}{4}\).}
\label{fig:rotate_full}
\end{figure}

\textit{Interpretation of attribute values for non-negative less distinct components}. Now, we consider data sample with lower \(a_1=0.7\) (i.e. less distinct) but components are still non-negative. Figure \ref{fig:rotate} middle shows that components are still non-negative around \(\theta\in[\pi/8,3\pi/8]\). Similar attribution of \(h_1\) and suppression of \(h_2\) are observed similarly although with lower magnitude around \(\theta\approx 0\). At \(\theta\approx \pi/4\), similar difficulty in distinguishing homogenous transformation is present, naturally. Further rotation to \(3\pi/8\) will give higher \(h_2\) as well. Figure \ref{fig:rotate_full} right shows similar behavior even for \(a_1\approx a_2\), though non-negative values are observed for rotations around \([-\pi/4,\pi/4]\). The sample is barely categorized as \(c=1\) since \(a_1>a_2\). However, the resulting attribution values still highlights the positive contribution \(x_1\), primarily through higher \(h_1\) attribution value, even though the magnitudes are lower compared to previous examples. 

\textit{Interpretation of attribute values for negative components}. Beyond the rotation range that yields non-negative components, we do see negative components \(x_i<0\) assigned highly negative \(h_i\) values. For example, Figure \ref{fig:rotate_full} left at \(\theta\approx \pi\) shows a rotation of the components to the negatives. In this formulation, negative attribution values are assigned to negative components naturally, because \(w*x\) starts with \(w_i=1\) and \(x_i<0\), as \(w_i\) is optimized, our example shows an instance where, indeed, we need higher \(w_i\), very negative \(h_1\). Recall the main interpretation. In the end, this high negative attribution is aimed at improving CO score. The large negative \(h_1\) component increases the likelihood of predicting \(c=1\); conversely, the relatively low \(h_2\) magnitude increases the same likelihood. Therefore, we do not immediately conclude that negative attribution values contribute ``negatively" to prediction, which is a term sometimes ambiguously used in XAI community. In practice, case by case review may be necessary.

\textit{More relevant works}. There have been many different papers on interpretable models not included here, particularly because they are not directly related to heatmap-based XAI. Regardless, some models interpretable in different ways include \cite{zhang2019unsupervised, tjoa2021two} and some other related to reinforcement learning have been described in \cite{tjoa2021self}.

\subsection{Zero CO scores and other scores}
Zero CO score might occur when when \(h\) yields uniform change to the output, i.e. \(DNN(x+h)=DNN(x)+c\) for some constant \(c\). This is obtained by simply plugging into the CO score formula. Special case may occur when \(h\) is constant over all pixels, especially when \(N(g(x+h))=N(g(x))\) for some intermediate normalization layer \(N\) and intermediate pre-normalized composition of layers \(g=g_k \circ g_{k-1} \circ \dots \circ g_1\).

Positive CO score indicates that the component \([s_{co}]_i\), where \(i\) corresponds to the groundtruth label, is increased by \(h\) at a greater magnitude than the average of all other components, which in turn means that the component \(DNN(x+h)_i\) is similarly increased at greater magnitude compared to the average of other components. Hence, the prediction favours component \(i\) relatively more, i.e. the probability of predicting the correct class is increased. Negative CO score is simply the reverse: probability of predicting the correct class has been reduced relatively. 
%%%%%%%%%%%%%%%%%%%%%%%%%%%%%%%%%%%%%%%%%%%%%%%%%%%%%%%%%%%%%%%%%%%%%%%%%%%%%%%
%%%%%%%%%%%%%%%%%%%%%%%%%%%%%%%%%%%%%%%%%%%%%%%%%%%%%%%%%%%%%%%%%%%%%%%%%%%%%%

\subsection{More Boxplots of CO Scores.}
\begin{figure*}[]
\centering
\includegraphics[width=1\columnwidth]{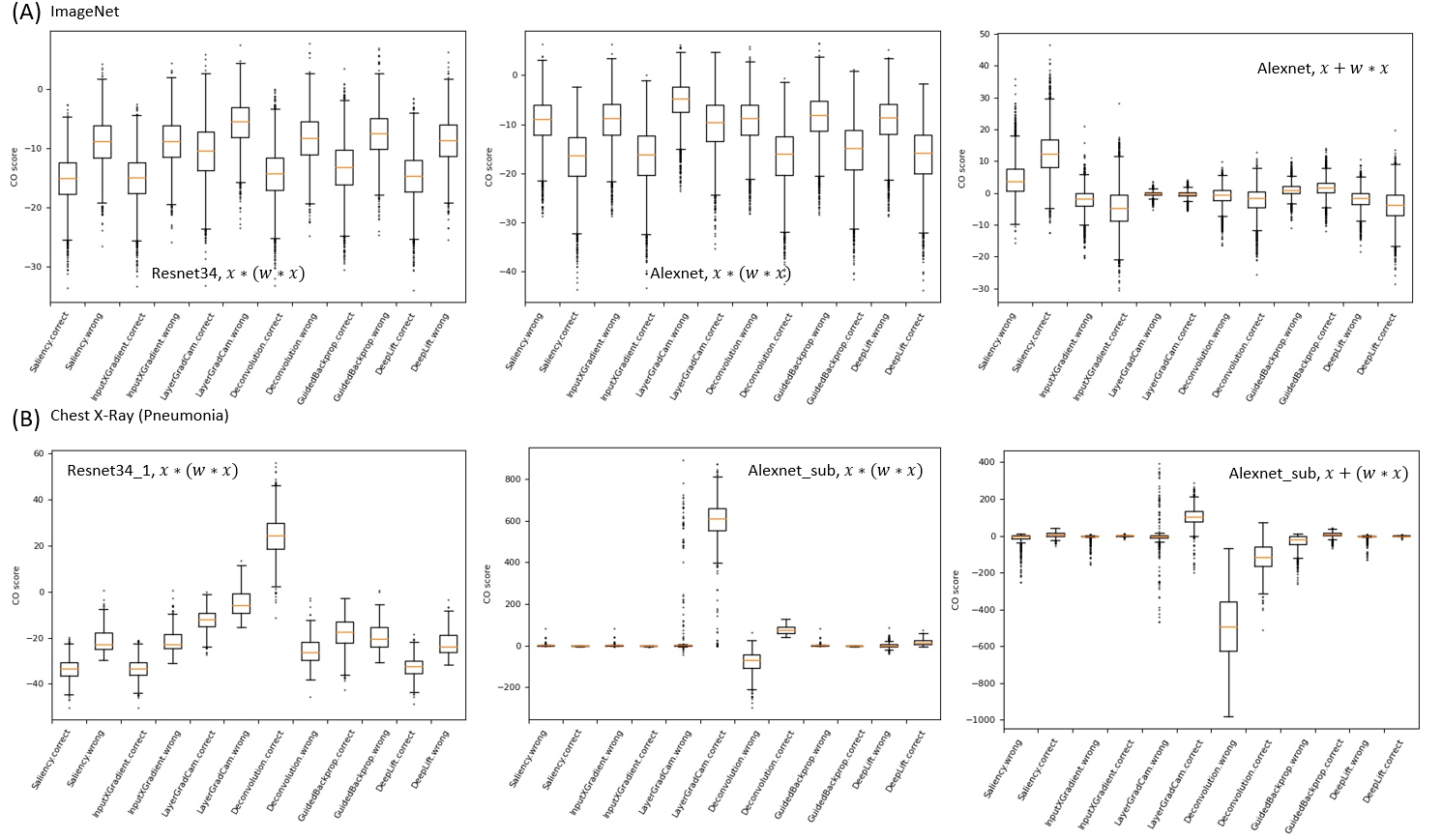} 
\caption{Boxplots of CO scores for existing XAI methods, including another GAX implementation \(x*h=x*(w*x)\)}.
\label{fig:more_boxplots}
\end{figure*}

Figure \ref{fig:more_boxplots} shows more variations of CO scores in our experiments, similar to the ones shown in the main text. Some scores clearly demonstrate distinct gaps in CO scores between the correct and wrong predictions.

\begin{figure*}[]
\centering
\includegraphics[width=1\columnwidth]{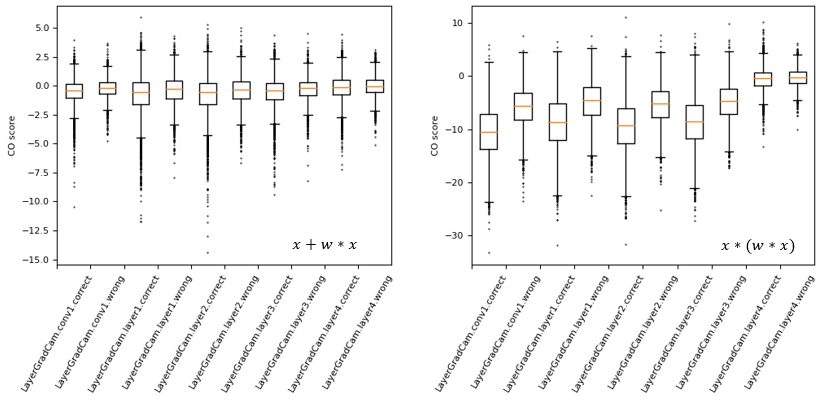} 
\caption{Boxplots of CO scores for heatmaps from Layer GradCAM for ResNet34 and ImageNet dataset. CO scores of heatmaps generated from different layers (and resized accordingly) are shown.}.
\label{fig:layer_comparison_boxplots}
\end{figure*}

From Figure \ref{fig:layer_comparison_boxplots}, AX process is applied to the heatmaps generated in different layers of ResNet34. We expect higher improvement of CO scores for AX process using heatmaps from deeper layers that are known to detect more features. We do observe a difference in \(x*h\) AX process, but not in \(x+h\) for Layer Grad CAM.

We have re-run the experiments in the main text with version 2 of the codes. The only difference is the use of pytorch version 2; the results are virtually the same.

\begin{figure*}[]
\centering
\includegraphics[width=1\columnwidth]{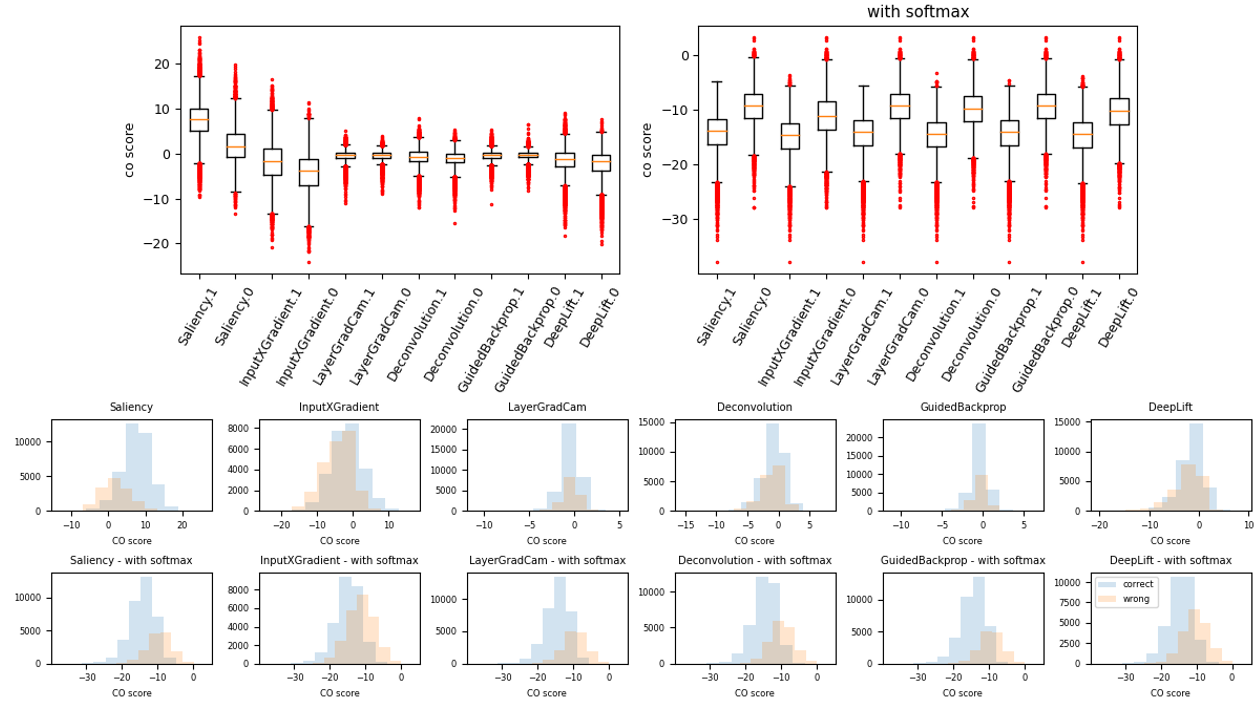} 
\caption{Boxplots of CO scores and histograms for ImageNet with Resnet34, rerun using pytorch version 2. 1/0 denote correct/wrong respectively.}
\label{fig:}
\end{figure*}

\begin{figure*}[]
\centering
\includegraphics[width=1\columnwidth]{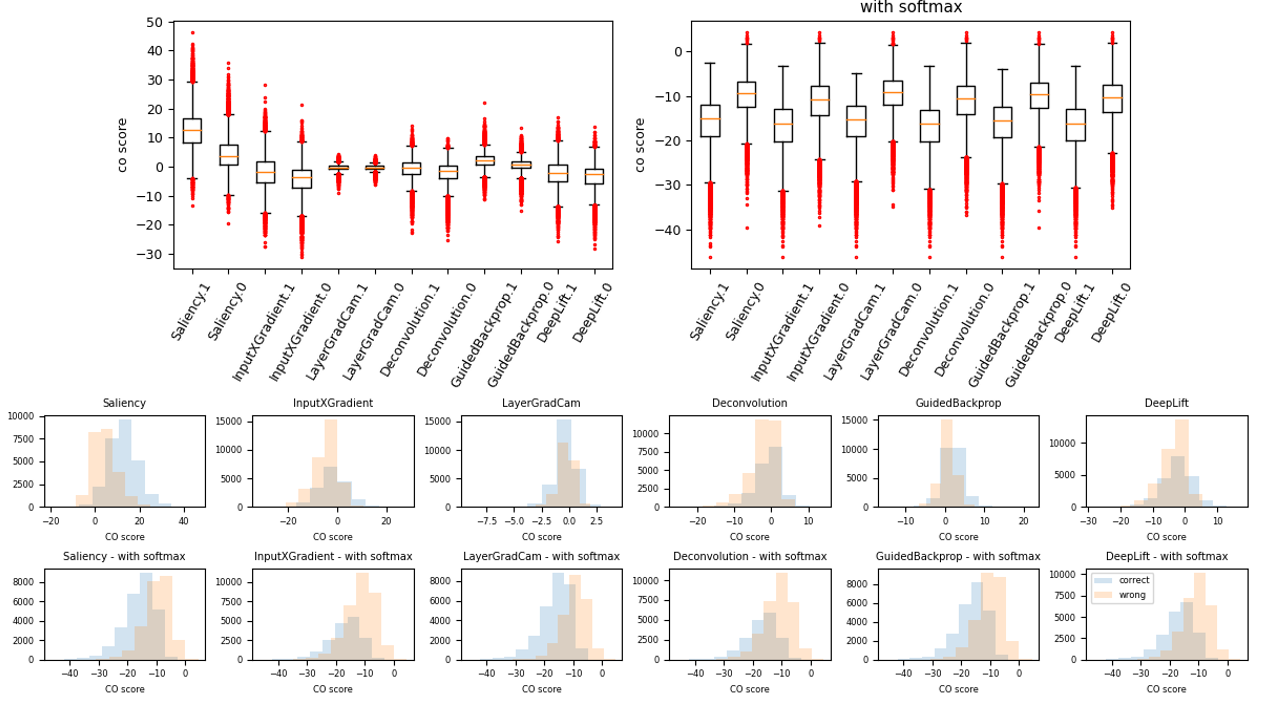} 
\caption{Boxplots of CO scores and histograms for ImageNet with Alexnet, rerun using pytorch version 2.}
\label{fig:}
\end{figure*}

\begin{figure*}[]
\centering
\includegraphics[width=1\columnwidth]{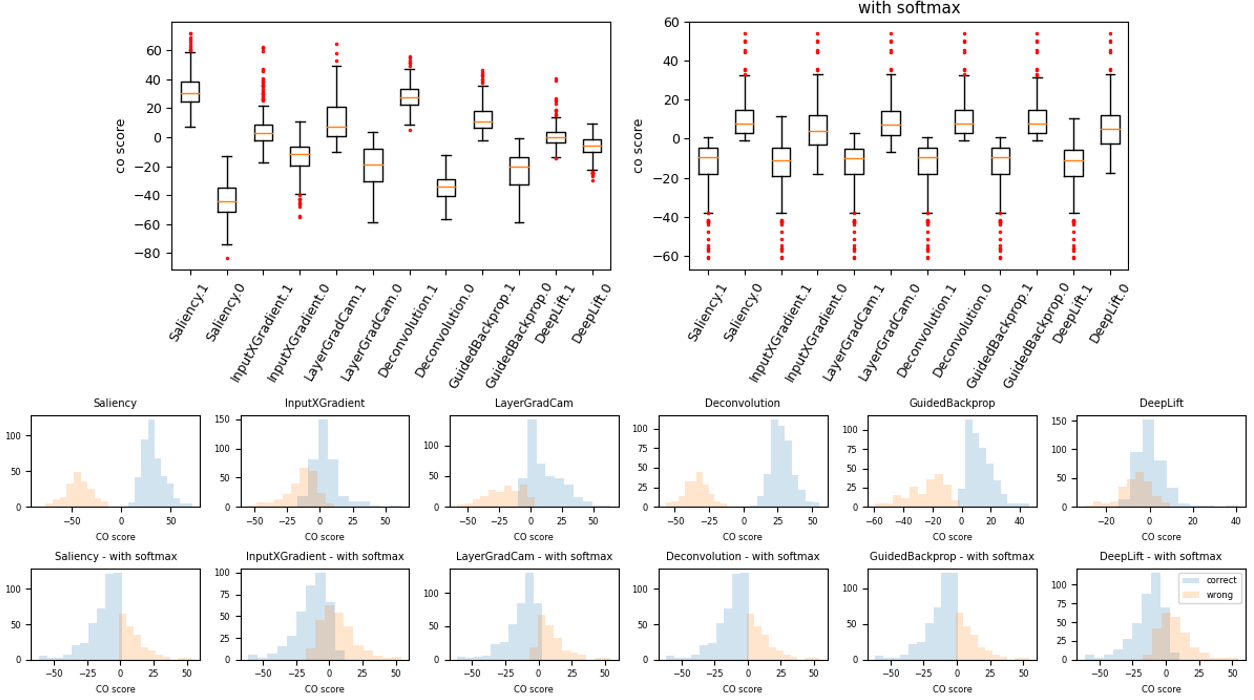} 
\caption{Boxplots of CO scores and histograms for chest X-Ray dataset for pneumonia classification with Resnet34, rerun using pytorch version 2.}
\label{fig:}
\end{figure*}

\begin{figure*}[]
\centering
\includegraphics[width=1\columnwidth]{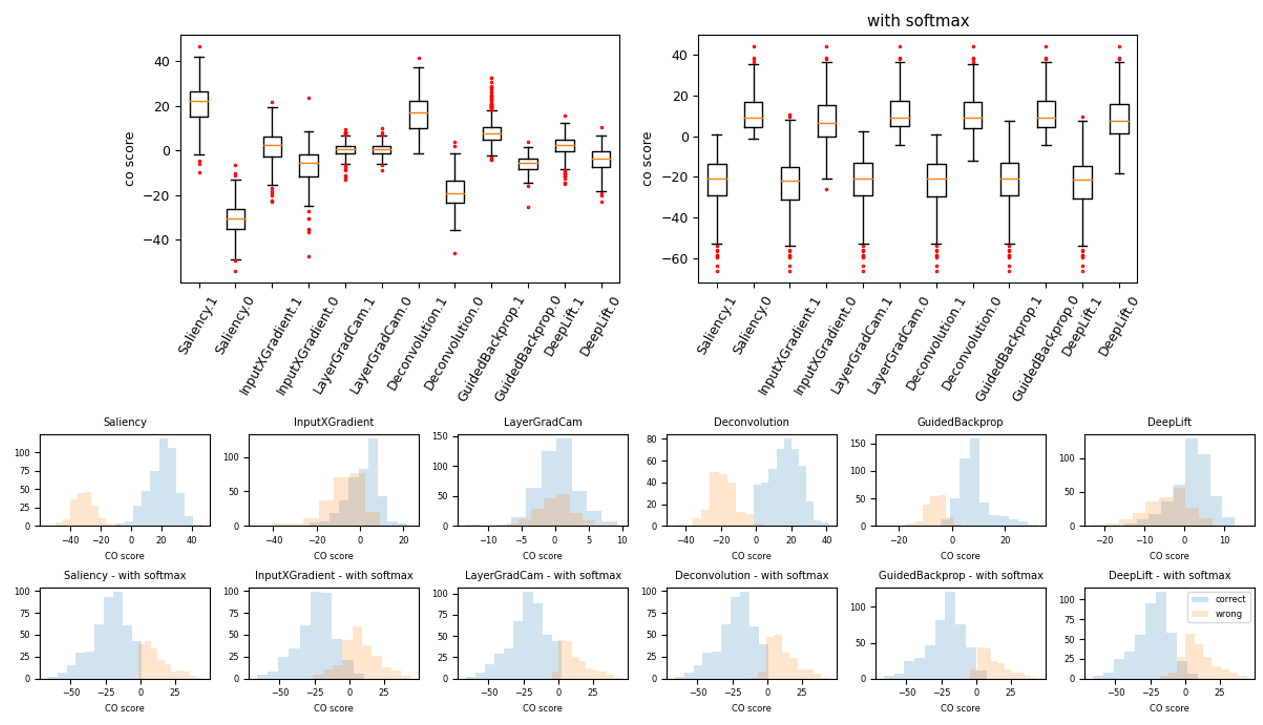} 
\caption{Boxplots of CO scores and histograms for chest X-Ray dataset for pneumonia classification with Resnet34, trained sub-optimally, rerun using pytorch version 2.}
\label{fig:}
\end{figure*}

\begin{figure*}[]
\centering
\includegraphics[width=1\columnwidth]{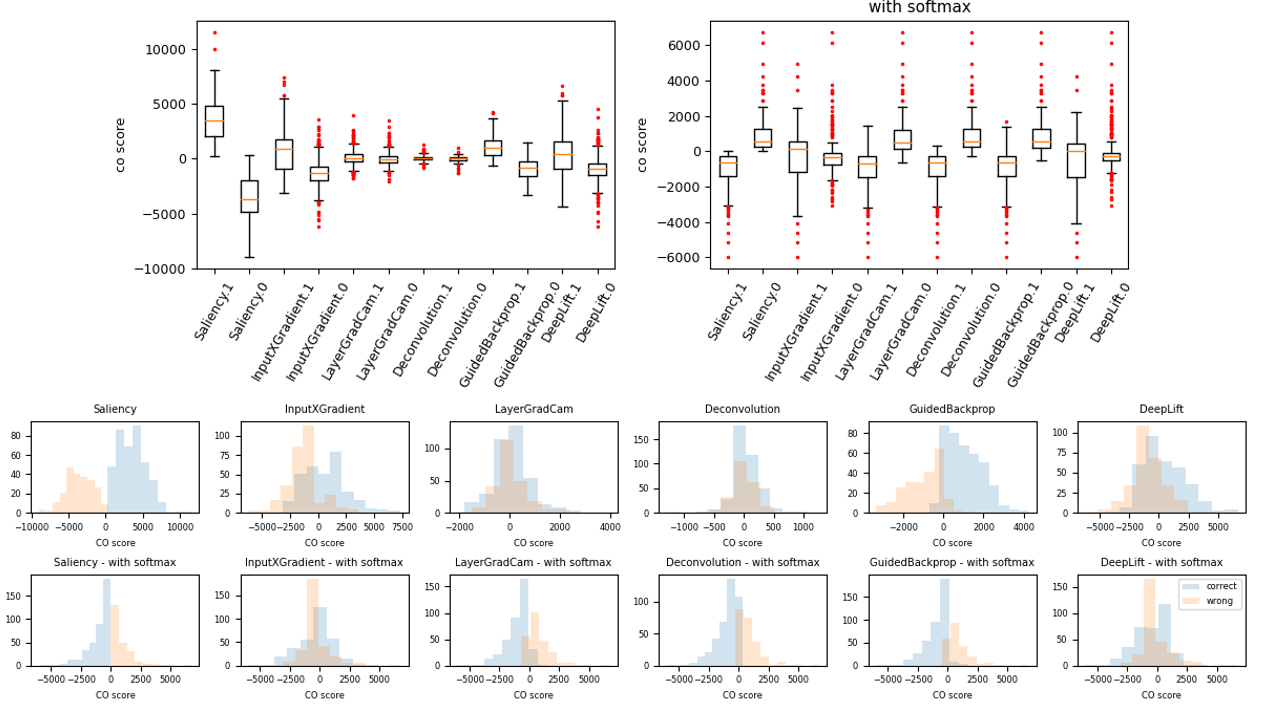} 
\caption{Boxplots of CO scores and histograms for chest X-Ray dataset for pneumonia classification with Alexnet rerun using pytorch version 2.}
\label{fig:}
\end{figure*}

\subsection{Experiments and Results on more datasets}
Here, we describe similar experiments on 3 other datasets (1) COVID-19 Radiography Database \cite{chowdhury2020can, rahman2021exploring} \footnote{\url{https://www.kaggle.com/datasets/tawsifurrahman/covid19-radiography-database}} (2) credit card fraud detection\footnote{https://www.kaggle.com/datasets/mlg-ulb/creditcardfraud} \cite{dal2015calibrating} (3) dry bean classification\footnote{\url{https://www.kaggle.com/datasets/muratkokludataset/dry-bean-dataset}} \cite{koklu2020multiclass}. All details can be found in our github; we will describe them briefly here. Note: we also tested CO scores on output with softmax, as shown in the respective CO score distribution plots. We initially expected the CO scores distribution with softmax to be much less distinct since softmax squashed the magnitudes, causing output channels to have more similar values. Interestingly, while this is true for some cases, it appears that some softmax versions produce clear distribution gaps as well.\\

\noindent\textbf{COVID-19 Radiography Database}\\
We model COVID-19 Radiography Database (henceforth COVID) classification problem with CXCMultiSPA, a small convolutional neural network with Spatial Product Attention (SPA) \cite{Tjoa2020QuantifyingEO}. The choice of model is almost completely arbitrary, and our intention is to use such a small network (only 50565 parameters) to demonstrate that reasonably high accuracy can still be attained.

\begin{figure*}[]
\centering
\includegraphics[width=1\columnwidth]{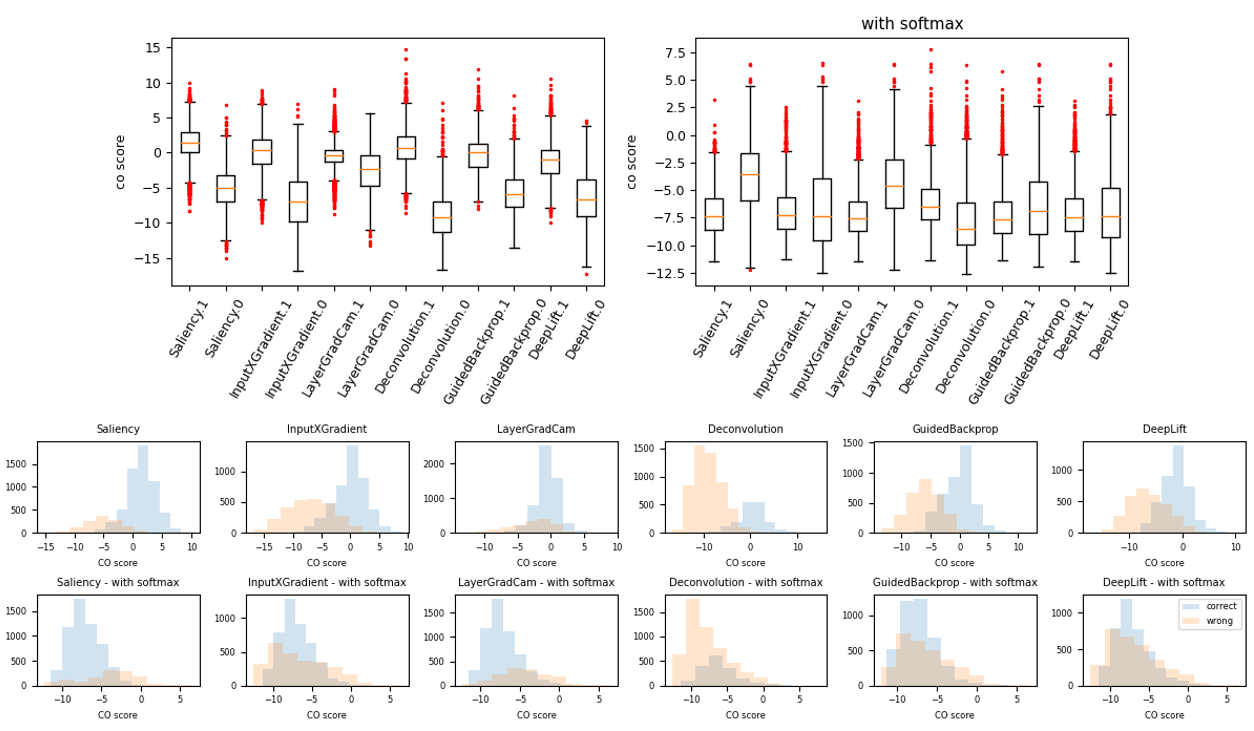} 
\caption{Boxplots of CO scores and histograms for COVID-19 Radiography Database with CXCMultiSPA. 1/0 denote correct/wrong respectively. }
\label{fig:boxplot_chestxray_covid}
\end{figure*}

The task is to classify chest X-ray images and to predict if the patient is normal or suffers from COVID-19, lung opacity or viral pneumonia (hence there are four classes). The train, validation and test splits are constructed in a deterministic way as the following. For each folder (corresponding to one of the four classes), iterate through all data samples indexed by \(i\), \(i=0\) to \(i=n-1\) where \(n\) is the total number of images in the folder. Let \(i'=i\ mod\ 3\), if \(i'=0,1,2\), then the sample is put into the train, val and test folders respectively. The training and validation settings are as the following. Batch size is 16, images are resized to (256,256), Adam optimizer is used with learning rate of 0.001 and the training is allowed to run for 256 epochs or until validation accuracy of 0.85 is achieved. The final test accuracy is 0.865. CO scores distribution are shown in Figure \ref{fig:boxplot_chestxray_covid}. The trend is similar to our previous results: there are distributional gaps between the CO scores of correct and wrong predictions. Softmax version appears squashed i.e. some gaps are less obvious, and the difference between the scores of correct and wrong predictions is less distinct.\\

\noindent\textbf{Creditcard Fraud Detection}\\
We model creditcard fraud detection with ccfFPA, a model based on Feature Product Attention (FPA). FPA is practically a 1D convolutional version of SPA. This choice of model is also arbitrary, and it is also a small model with only 506 parameters. Here, we use the 28 PCA features available in the dataset (V1-V28) to predict whether the given transaction is genuine or fraudulent.

Since this dataset is highly imbalanced, only 492 frauds (positive) out of 284,807 samples, we use a slightly different augmentation technique, which we call the \textit{cross projective augmentation}. Let \(D_F\) denote the set of samples with positive label (fraud) and \(D_N\) the set of samples with negative label (normal transaction). The training dataset is constructed as the following. First, we arbitrarily collect \(T\subset D_N\) such that \(|T|=c\times|D_F|\), where \(c=5\) is called the \textit{cross factor}. Then, for every sample \(x\in D_F\), pick \(t_i\in T\) for \(i=1,2,\dots, c\) and let \(T_x=\{t_i+d(x-t_i)|i=1,\dots,c\}\) where \(d=0.95\) so that \(T_x\) and \(T_{x'}\) do not intersect for any \(x,x'\in D_F\). The training dataset is then \(T\cup \Big(\bigcup_x T_x\Big)\). In essence, we use some synthetic samples near the test samples as the data for training. Validation dataset is constructed similarly.

\begin{figure*}[]
\centering
\includegraphics[width=1\columnwidth]{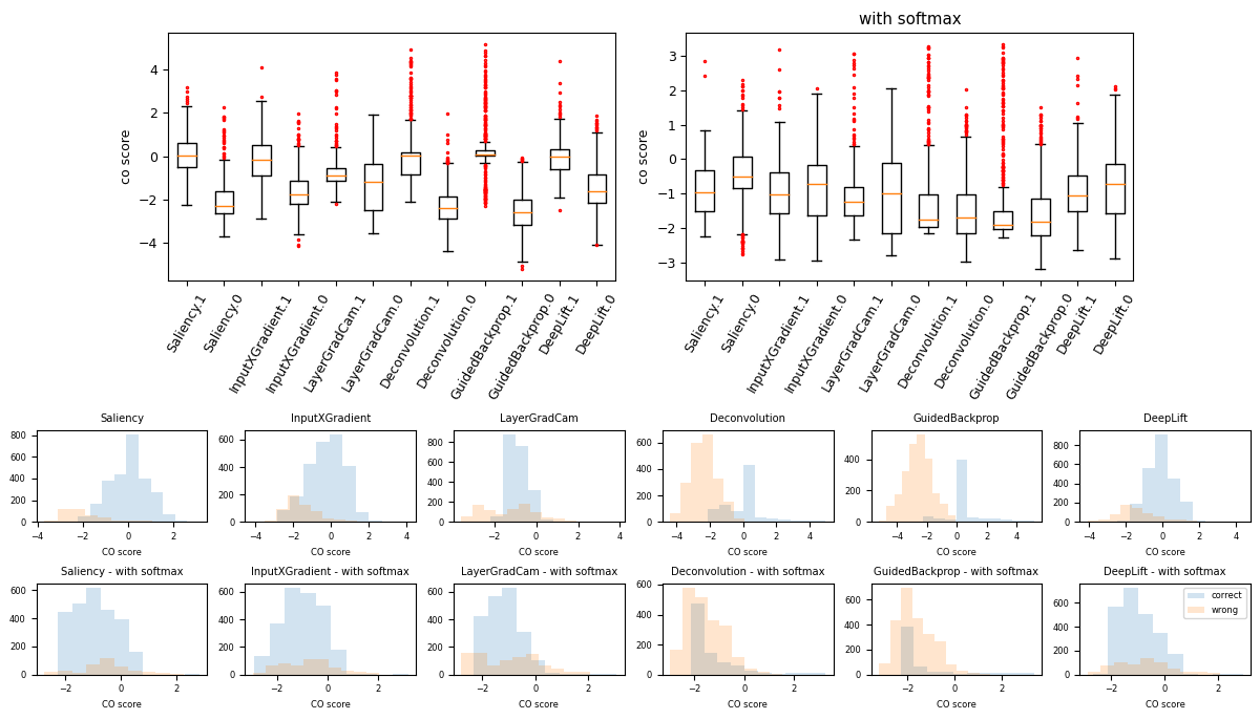} 
\caption{Boxplots of CO scores and histograms for creditcard fraud dataset with ccfFPA model. 1/0 denote correct/wrong respectively. }
\label{fig:boxplot_ccf}
\end{figure*}

The training and validation settings are as the following. Adam optimizer is used with learning rate of 0.001 and the training is allowed to run for 256 epochs or until validation accuracy of 0.9 is achieved. The final test accuracy is 0.895. CO scores distribution are shown in Figure \ref{fig:boxplot_ccf}. The trend is similar to our previous results.\\

\noindent\textbf{Drybean dataset}\\
We model drybean classification dataset with an FPA-based model as well, namely drybeanFPA. This choice of model is also arbitrary, and it is also a small model with only 2121 parameters. The drybean dataset consists of samples with 1D vector of features, and there are 7 classes of beans.

The data is reasonably balanced, so we construct the train, validation and test dataset like how we did with COVID dataset. For every class of bean Bombay, Seker, Barbunya, Dermason, Cali, Horoz and Sira, we go through each sample indexed with \(i=0\) to \(i=n-1\) where \(n\) is the number of samples in that class. Then let \(i'=i\ mod\ 3\). If \(i'=0,1,2\), then the sample is put into the train, val and test folders respectively.

\begin{figure*}[]
\centering
\includegraphics[width=1\columnwidth]{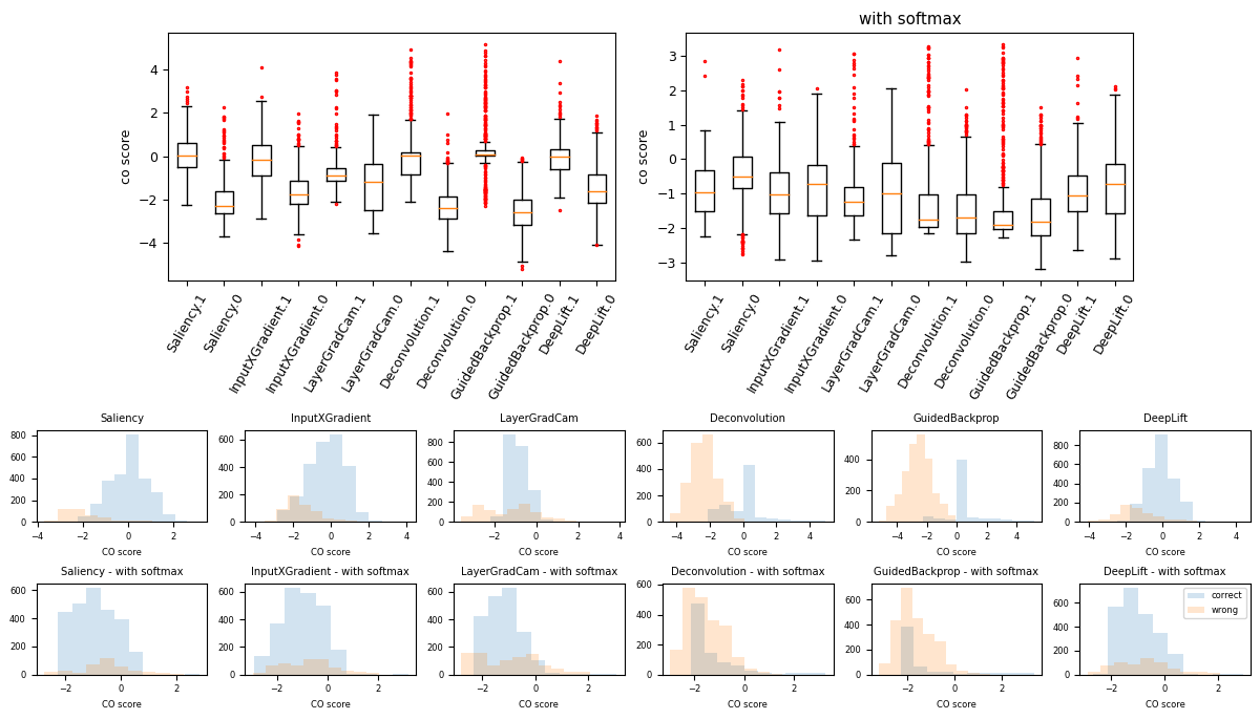} 
\caption{Boxplots of CO scores and histograms for dry bean dataset with drybeanFPA model. 1/0 denote correct/wrong respectively. }
\label{fig:boxplot_drybean}
\end{figure*}

The training and validation settings are as the following. Adam optimizer is used with learning rate of 0.001 and the training is allowed to run for 256 epochs or until validation accuracy of 0.85 is achieved. The final test accuracy is 0.868. CO scores distribution are shown in Figure \ref{fig:boxplot_drybean}. The trend is similar to our previous results

\subsection{Accuracy comparison}
We compare the classification result AX process applied on all of the above XAI methods, datasets and architectures in table \ref{table:allacc}. \textit{Baseline} refers to the original accuracy without AX process. Unfortunately, improvement has been limited as mentioned in the main text.

\begin{table}[]
\caption{Comparison of validation or testing accuracies before and after augmentative explanation process \(x+h\). IMN: ImageNet Pneu: chest x-ray for pneumonia detection. COVID: COVID-19 Radiography Database. CCF: credit card fraud detection . DB: dry bean classification. IXG: InputXGradient. LGC: LayerGradCam. GBP: GuidedBackprop.	 }\label{table:allacc}
\begin{tabular*}{\tblwidth}{cccccccc}
\toprule
& Baseline	& Saliency	& IXG	 & LGC & Deconvolution &	GBP &	DeepLift \\
	\midrule
IMN, Resnet34 &	0.722 &	0.715 &	0.526 &	0.703	 & 0.659 & 	0.706 & 	0.610 \\
IMG, Alexnet	 & 0.551 & 	0.53	0 & 0.347 &	0.542 & 	0.364 & 	0.484 & 	0.407 \\ 
Pneu, Resnet34 & 	0.691 &	0.691&	0.601&	0.671&	0.691&	0.689	&0.619\\
Pneu, \texttt{Resnet34\_sub}&	0.708	&0.708	&0.643&	\color{cyan}{0.713} &	0.700&	\color{cyan}{0.712}&	0.668\\
Pneu, Alexnet	& 0.628 &	0.628	& 0.51	& 0.606	& \color{cyan}{0.636}	& \color{cyan}{0.635}	& 0.546 \\
COVID, CXCMultiSPA &	0.865 &	0.821 &	0.624 &	0.784 &	0.245 	& 0.578 &	0.562 \\
CCF, ccfFPA	& 0.895 &	0.872&	0.813	&0.692&	0.282 &	0.188 &	0.785\\
DB, drybeanFPA &	0.877	&0.868	&0.708	&0.646&	0.817	& 0.777	&0.712\\
\bottomrule
\end{tabular*}
\end{table}

\subsection{More Considerations, Limitations and Future Works}

\textit{Different GAX and empirical choices in implementation}. Parts of the implementations, such as the initialization of \(w\) to 1.0, are nearly arbitrary, though it is the first choice made from the 2D example that happens to work. Different implementations come with various trade-offs. Most notably, the choice of learning rate \(0.1\) is manually chosen for its reliable and fast convergence, although convergence is attainable for smaller learning rate like \(0.001\) after longer iterations. However, we need to include more practical considerations. For example, saving heatmaps iteration by iteration will generally consume around 5-12 MB of memory for current choices. Longer optimization iterations may quickly cause a blow-up, and there is no known fixed number of iterations needed to achieve convergence to the target CO score. Saving heatmaps at certain CO scores milestones can be considered, though we might miss out on important heatmap changes in between. Parameter selection process is thus not straightforward. For practical purposes, learning rates can be tested in order of ten, \(10^n\), and other parameters can be tested until a choice is found where each optimization process converges at a rate fast enough for nearly instantaneous, quick diagnosis. Other choices of optimizers with different parameters combination can be explored as well, though we have yet to see dramatic changes. 

\textit{GAX, different DNN architectures and different datasets}. Comparisons are tricky, since different architectures might behave differently at their FC end. For example, for Saliency method on ImageNet, Alexnet's boxplot of CO scores in appendix Figure \ref{fig:more_boxplots}(A) right (AX process \(x+w*x\)) shows a wider range of CO scores than that of Resnet34 in Figure \ref{fig:more_boxplots}. Comparison of CO scores on Chest X-Ray dataset shows even larger variability. Furthermore, recall that we illustrated using the 2D example the reason we avoid sigmoid function: suppressed change in the CO score due to its \textit{asymptotic} part. From here, the ideal vision is to develop a model that scales with CO score in not only a computationally relevant way, but also in a human relevant way: we want a model that increases predictive probability when the heatmap highlights exactly the correct localization of the objects or highly relevant features related to the objects. This is a tall effort, particularly because explanations are highly context dependent. Transparent and trustworthy applications of DNN may benefit from the combined improvements in humanly understandable context and computationally relevance attributions built around that context.

\end{document}